\documentclass[10pt,twocolumn]{article} 
\usepackage{simpleConference}
\usepackage{times}
\usepackage{graphicx}
\usepackage{amssymb}
\usepackage{url,hyperref}
\usepackage{CJK}
\usepackage{makecell}
\usepackage{natbib}
\usepackage{multibib}
\setcitestyle{authoryear}
\usepackage{mdframed}
\usepackage{listings}
\usepackage{multirow}
\usepackage{verbatim}
\usepackage{makecell}
\usepackage{setspace}
\usepackage[hang,flushmargin]{footmisc} 

\usepackage[T1]{fontenc}

\usepackage{lipsum}
\newcommand\blfootnote[1]{%
  \begingroup
  \renewcommand\thefootnote{}\footnote{#1}%
  \addtocounter{footnote}{-1}%
  \endgroup
}

\title{Exploring the Impact of Instruction Data Scaling on Large Language Models: An Empirical Study on Real-World Use Cases}

\author{Yunjie Ji\textsuperscript{\#}, Yong Deng\textsuperscript{\#}, Yan Gong, Yiping Peng, Qiang Niu, Lei Zhang, Baochang Ma\textsuperscript{*}, Xiangang Li  \\
Beike Inc., Beijing, China  \\
\texttt{\{jiyunjie001,dengyong013,gongyan013,pengyiping001, }  \\
  \texttt{niuqiang002,zhanglei252,mabaochang001,lixiangang002\}@ke.com}}
  
\begin{document}
\maketitle
\begin{abstract}
The success of ChatGPT has recently attracted numerous efforts to replicate it, with instruction-tuning strategies being a key factor in achieving remarkable results. 
Instruction-tuning not only significantly enhances the model's performance and generalization but also makes the model's generated results more consistent with human speech patterns. 
However current research rarely studies the impact of different amounts of instruction data on model performance, especially in the real-world use cases.
In this paper we explore the performance of large language models based on instruction tuning across different scales of instruction data. 
An evaluation dataset consisting of 12 major online use cases is constructed in the experiment. 
With Bloomz-7B1-mt as the base model, the results show that 1) merely increasing the amount of instruction data leads to continuous improvement in tasks such as open-ended generation, 2) in tasks such as math and code, the model performance curve remains quite flat while increasing data size.
We further analyze the possible causes of these phenomena and propose potential future research directions such as effectively selecting high-quality training data, scaling base models and training methods specialized for hard tasks.
We will release our training and  evaluation datasets, as well as model checkpoints\textsuperscript{1}\blfootnote{
\textsuperscript{\#}Equal contribution \\
      \textsuperscript{*}Corresponding author \\
\textsuperscript{1}https://github.com/LianjiaTech/BELLE
}. 
\end{abstract}


\section{Introduction}
The purpose of instruction-tuning \cite{FinetunedLanguageModels2021, MultitaskPromptedTraining2021, ScalingInstructionFinetunedLanguage2022, TrainingLanguageModels2022} is to enable models to understand and correctly respond to various human instructions. 
The key is to guide the model to comprehend task requirements by concatenating a text describing the task as an instruction before the input text. 
Unlike fine-tuning a model to solve a specific NLP task, instruction-tuning aims to improve the model's generalization capability on unseen tasks, which is achieved by dealing with all tasks in a way of generation and training with various types of instructions.

Recently, Models trained with human feedback \cite{TrainingLanguageModels2022, ConstitutionalAIHarmlessness2022, FineTuningLanguageModels2020, LearningSummarizeHuman2022, RedTeamingLanguage, WebGPTBrowserassistedQuestionanswering2022, PretrainingLanguageModels2023} (especially ChatGPT and GPT-4) have attracted significant attention from researchers in the field of artificial intelligence because it can generate high-quality responses to human input and even self-correct previous errors based on subsequent dialogues. 
Instruction-tuning strategy is one of the  key factors in achieving remarkable results with ChatGPT. 
To replicate ChatGPT, research community \cite{alpaca, openchatkit} focuses on obtaining a capable instruction-following model primarily by fine-tuning large language model on diverse and high-quality instruction datasets.

However the impact of instruction data size has not been well explored, especially for evaluation with typical use cases coming from online ChatGPT users. 
\cite{HolisticEvaluationLanguage2022, ChatGPTGeneralPurposeNatural2023, ye2023comprehensive, bang2023multitask, srivastava2022beyond, suzgun2022challenging} evaluated available large language models, but didn't pay attention to the influence of training strategies.
Meanwhile, most evaluations concentrated on conventional NLP tasks and were performed using datasets in English. 
 To fill these gaps, we construct a diverse and high-quality Chinese instruction training and evaluation dataset, and conduct extensive experiments to analyze the performance of models on different scales of instruction data. 
 Finally we obtain the following important experimental results:
 
 \begin{itemize}
\item 
In tasks like brainstorming and translation, a dataset of 2 million samples, or even less, can enable the model to achieve satisfactory performance.
\item 
Increasing data size still leads to performance improvement in tasks like open QA and extraction, indicating that the bottleneck has not been reached. But the potential for improvement may be limited.
\item
The model's performance is still poor for math and code, and increasing data size no longer brings about performance improvement. 
This indicates some future research directions such as effectively selecting high-quality training data, scaling base models in terms of parameters and basic abilities, and training methods specialized for tasks like math and code.
\end{itemize}

In summary, we conduct experiments on the impact of training data size on the performance of instruction-following models, and obtain several preliminary conclusions, which provide directions for future work. 
At the same time, we will open source our training and evaluation data, as well as the checkpoints of our models.

\section{Related Work}

\subsection{Large language models}
Transformer-based language models, especially the generative large language models have greatly advanced the development of Natural Language Processing \cite{vaswani2017attention, devlin2018bert, lan2019albert, yang2019xlnet, dong2019unified, clark2020electra, raffel2020exploring, LanguageModelsAre2020, OPTOpenPretrained2022, PaLMScalingLanguage2022, GPTNeoX20BOpenSourceAutoregressive2022, TrainingComputeOptimalLarge2022, glaese2022improving, srivastava2022beyond}.
The GPT (Generative Pre-trained Transformer) family of models is a remarkable instance, and its ability to comprehend and adhere to human instructions has been enhanced by RLHF \cite{TrainingLanguageModels2022, ConstitutionalAIHarmlessness2022, FineTuningLanguageModels2020, LearningSummarizeHuman2022, RedTeamingLanguage, WebGPTBrowserassistedQuestionanswering2022, PretrainingLanguageModels2023} in ChatGPT. 
As a result, ChatGPT has evolved from being a basic NLP task solver to a complete natural language assistant that can perform duties such as generating conversations and detecting errors in a piece of code.

\subsection{Instruction tuning}
Instruction-tuning is a new trend emerging from \cite{FinetunedLanguageModels2021, MultitaskPromptedTraining2021, mishra2021cross}, which seeks to improve the performance of language models by teaching them to follow natural language.
By formatting all tasks into natural language, generative language models are capable of dealing with almost all of NLP tasks. 
Early research focused on instruction tuning a general NLP task solver, and there is a trend towards converting more and more NLP datasets into a unified dataset then conducting multi-task training \cite{xu2022zeroprompt, xie2022unifiedskg, wang2022super, khashabi2020unifiedqa, min2021metaicl, ye2021crossfit, liu2019multi, zhong2021adapting, ScalingInstructionFinetunedLanguage2022}.
However these models still struggle with understanding general human instructions especially in real-world use cases.
Until the emergence of training methods like RLHF \cite{TrainingLanguageModels2022, ConstitutionalAIHarmlessness2022, FineTuningLanguageModels2020, LearningSummarizeHuman2022},  models truly began to understand various human instructions and produce good responses.
Recently, research community has delivered great efforts in replicating ChatGPT \cite{alpaca, openchatkit}.
In their work, the amount of data and types of tasks vary greatly, and the impact of these factors on model performance has not been well explored. 


\subsection{Evaluation of LLMs}
There are many evaluations of large language models, such as OPT \cite{OPTOpenPretrained2022}, BLOOM \cite{BLOOM176BParameterOpenAccess2022}, GLM \cite{zeng2023glm-130b}, and GPT-3 \cite{LanguageModelsAre2020}, in various tasks. \cite{HolisticEvaluationLanguage2022} conducted a thorough evaluation of 30 large language models.
\cite{ChatGPTGeneralPurposeNatural2023} evaluated the performance of ChatGPT on various NLP tasks. 
\cite{ye2023comprehensive} compared the capabilities of GPT and GPT-3.5 series models. 
\cite{bang2023multitask} compared the reasoning, hallucination reduction, and interactivity abilities of ChatGPT in multiple languages and modalities.
However, these evaluations mainly focus on the performance of existing models and do not evaluate the performance of models under different scales of instruction data. 
Additionally, many evaluation data consist of  traditional NLP tasks, which differ from real-world human usage scenarios.
\cite{srivastava2022beyond} provided 204 tasks, which are believed to be beyond the capabilities of current large language models. 
\cite{suzgun2022challenging} selected the 23 most difficult tasks from BIG-Bench, forming BIG-Bench Hard (BBH).
Our proposed evaluation dataset is closer to real-world human usage scenarios and is dedicated to the Chinese community.

\begin{table}[t!]
\caption{The number of and average prompt length of each type of instructions.}
\small
\begin{center}
\begin{tabular}{c|c|c} 
\hline 
\textbf{Use case} & \textbf{\#Nums} & \textbf{Average prompt length}\\
\hline   
Math  & 200 & 49.15 \\
\hline
Code & 174 & 66.18 \\
\hline
COT & 197 & 23.92 \\
\hline
Classification & 200 & 54.75 \\
\hline
Extraction  & 194 & 73.89 \\
\hline
Open QA & 190 & 22.55 \\
\hline
Closed QA & 189 & 181.79 \\
\hline
Generation & 187 & 43.19 \\
\hline
Brainstorming & 190 & 22.03 \\
\hline
Rewrite & 200 & 53.51 \\
\hline
Translation & 147 & 37.28 \\
\hline
Summarization & 142 & 105.53 \\
\hline
\end{tabular}
\end{center}
\label{data_stat}
\small
\end{table}

\section{Method}
In this section, we will introduce the method of  obtaining  high-quality instruction tuning data, and the method of constructing diversified test instructions.
Same as our previous work \cite{ji2023exploring}, ChatGPT is also required to evaluate responses generated by instruction-following models.
The prompts are listed in Appendix \ref{eval_method}.
\subsection{Generate training data}
Manual annotation of high-quality instruction data requires significant resources.
Given the powerful in-context learning ability, large language models can generate a great number of diverse instruction data based on high-quality seed set \cite{wang2022self2}. 
In this paper, we adapt the same method as \cite{alpaca}.
We translate the open-source seed data provided by \cite{alpaca} into Chinese and modify some of the data that heavily involve Western culture and background knowledge to be more in line with Chinese cultural and background knowledge. 
Then, using these seed data as in-context examples, we require ChatGPT to generate more samples. 
\subsection{Generate evaluation data}
 We select a portion of data generated from ChatGPT for evaluation.
 Annotators were asked to correct ChatGPT's responses to obtain the golden responses for test instructions. 
 Our test instructions are classified to 12 types, covering the most common use cases for online users. 
 Table \ref{data_stat} shows the detailed information of these test instructions.
In addition, we plan to continue expanding our evaluation dataset, as more data leads to more reliable evaluation results.

\section{Experiments}

\begin{figure*}[t!]
	\centering
	\includegraphics[scale=0.343]{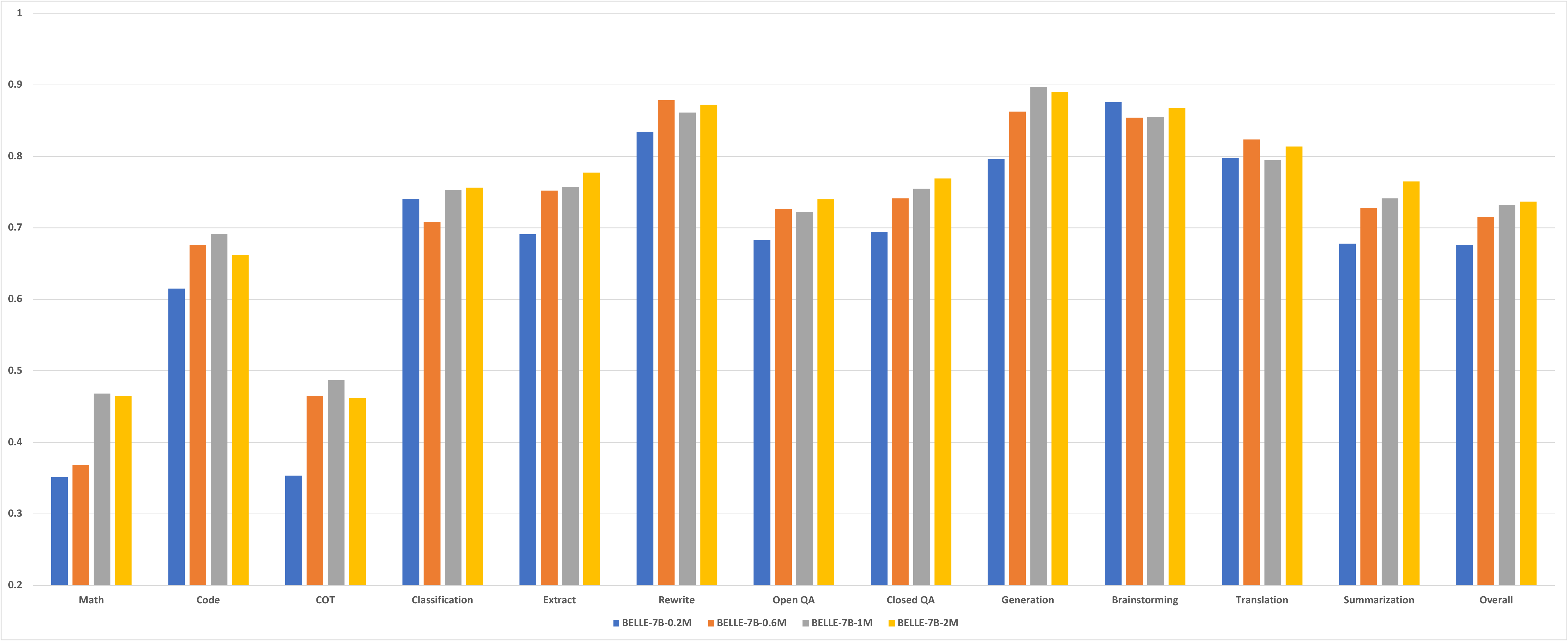}
	\caption{Scores on 12 types of instructions.}
\label{procedure}
\end{figure*}

\subsection{Instruction-following Models}
This paper focuses on model performance on Chinese text.
While LLAMA \cite{touvron2023llama}, OPT \cite{OPTOpenPretrained2022} and GPT-J \cite{gpt-j} have not been particularly optimized for  Chinese, we select Bloomz-7b1-mt\cite{BLOOM176BParameterOpenAccess2022, CrosslingualGeneralizationMultitask2022} as our base model, which has 7.1 billion parameters and is further finetuned on the xP3mt dataset based on Bloom-7b1.
As shown in Table \ref{models}, we train Bloomz-7b1-mt with 0.2 million, 0.6 million, 1 million and 2 million instruction examples to obtain BELLE-7B-0.2M, BELLE-7B-0.6M, BELLE-7B-1M and BELLE-7B-2M respectively.
In this paper we only explore the influence of data scale and leave the influence of model scale for future work.
We train these models with 64 batch sizes, 2 epochs, constant learning rate of 3e-6, weight decay of 0.001. 
For each instruction, our instruction-following models are required to generate responses once. 
Although the responses generated by the model for the same instruction may differ, we believe that such fluctuations have little impact on the experimental results.

\begin{table}[t!]
\caption{Instruction-following models trained from Bloomz-7B1-mt with different scales of instruction data. We name these series models as BELLEs which refers to Bloom Enhanced Large Language model Engines.}
\small
\begin{center}
\begin{tabular}{c|c} 
\hline 
\textbf{Datasize} & \textbf{Instruction-following model} \\
\hline   
200,000  & BELLE-7B-0.2M \\
\hline
600,000  & BELLE-7B-0.6M \\
\hline
1,000,000  & BELLE-7B-1M \\
\hline
2,000,000  & BELLE-7B-2M \\
\hline
\end{tabular}
\end{center}
\label{models}
\small
\end{table}

\subsection{Metrics}
As mentioned in \ref{eval_method}, ChatGPT is asked to evaluate responses generated by instruction-following models.
For all instructions, ChatGPT gives a score between 0 and 1, where score 0 is the worst and score 1 is the best. 
For each type of instruction, we calculate the model's average score on the test examples. 
Additionally, considering the volatility of ChatGPT's generations, each model response is evaluated three times and the scores are averaged. 
It is worth noting that we don't adopt self-consistency \cite{wang2022self2} because many types of instructions in our test set do not have a unique standard answer.
Evaluation is achieved by invoking gpt-3.5-turbo API at the time of March 25, 2023.

\subsection{Analysis}
For the overall score, as the amount of data increases, the model's performance improves continuously, while such continuous improvement is not always expectable across all types of instructions. 
At the same time, we find that the model has already achieved a good performance with only 200k training examples.

\noindent \textbf{Math, Code and COT } \quad 
For Math, Code, and COT instructions, the model's performance is poor with 200 thousand training examples.
After increasing the number of training examples to 1 million, the model's performance improves, then it becomes difficult to further improve the performance, and it is far from a satisfactory level. 
There may be two reasons for this: 1) the quality of these three types of training data is poor, so the performance improvement is suppressed by erroneous training data as the amount of data increases. 2) the model size is not large enough to achieve the emergence of abilities, so it cannot further improve on these three types of instructions which require reasoning abilities.

\noindent \textbf{Extraction, Classification, Closed QA and Summarization} \quad 
For instructions of extraction, classification, closed QA, and Summarization, which are common NLP tasks, increasing the amount of training data can continuously bring about performance improvement.
This indicates that we can still obtain further performance improvement by simply increasing training examples in future training plans. 
However, it is important to pay attention to whether increasing the proportion of these types of data will cause the performance decline on other types of instructions.

\noindent \textbf{Open QA} \quad
For Open QA, the model's performance is continuously improved as the amount of data increases. 
Solving this task requires parametric knowledge of the model, so we can conclude that increasing the amount of training data enables the model to produce factual answers better and reduce hallucinations.

\noindent \textbf{Translation} \quad
In the translation task, Belle-7b-0.2m has achieved good performance, indicating that the model's translation ability may come from the multilingual ability of Bloomz-7b1-mt. 

\noindent \textbf{Rewrite} \quad
In the rewrite task, the model is required to correct grammar errors or paraphrase the original text to make it more smooth and concise. 
This type of task is relatively simple, and the model performs well with only 600 thousand training examples, so we can focus on other tasks in the future. 

\noindent \textbf{Generation} \quad
In the generation task (e.g. generating an article on a certain topic, writing an email), increasing the data size from 200 thousand to 1 million results in a significant improvement in performance, after which the performance plateaus. 

\noindent \textbf{Brainstorming} \quad
In the brainstorming task, a dataset of 200 thousand  proved to be the optimal size for the model's performance. 
This may be due to the fact that responses to this type of instructions are diverse and lack clear standards for judging response quality, causing ChatGPT tends to give higher scores when scoring. 
It also indicates that large language models are good at responding to this type of instructions.

In summary, for translation, rewrite, generation, and brainstorming tasks, a data size of 2 million or even less can enable the model to perform well. 
For extraction, classification, closed QA, and summarization tasks, the model's performance can continue to improve with the increase of data size, indicating that we can still improve the model's performance through  simply increasing training data size. 
But the potential for improvement may be limited.
The model performance is still poor on math, code and COT instructions, and further exploration is needed in data quality, model scale, and training strategies.

\section{Conclusion and Future work}
In this paper, we evaluate the impact of different amounts of instruction data on model performance. 
We find that hundreds of thousands of training examples can achieve good results on translation, rewrite, generation, and brainstorming tasks. 
Increasing data size still leads to performance improvement in tasks such as extraction, classification, closed QA, and summarization, indicating that the bottleneck has not been reached. 
However, in tasks such as math, code and COT, the model performance is poor and increasing data size no longer brings about performance improvement. 

The above findings have pointed out three directions for our future work.
Firstly, we will continue to explore the limits of increasing the amount of data in extraction, classification, closed QA, and summarization tasks. 
Secondly, we will improve the quality of training data to further enhance model performance, especially in math, code, and COT where the training data generated by ChatGPT is of low quality. 
Additionally, effectively selecting high-quality data  is also worth investigating. 
Lastly, we will evaluate the impact of base models on performance, including the number of model parameters and base abilities of pre-trained language models.

\bibliographystyle{unsrtnat}
\bibliography{mainTemplatePDF}

\section{Appendix A}
\subsection{Prompt ChatGPT as an evaluator}
\label{eval_method}
 Our previous work \cite{ji2023exploring} has demonstrated that ChatGPT's ranking preferences are consistent with human to a certain extent.
 So in this paper, we treat ChatGPT as an annotator as well to evaluate the responses generated by instruction-following models. 
 Table \ref{eval_prompts} lists the prompts we used for different types of instructions. 

\begin{table*}[t!]
\caption{Prompts which are designed to require ChatGPT to evaluate instruction-following models.}
\small
\begin{center}
\resizebox{\textwidth}{!}{
\begin{tabular}{c|c} 
\hline 
\textbf{Use case} & \textbf{Prompt}  \\
\hline   
Math  & \makecell[l]{\begin{CJK*}{UTF8}{gkai} 你是一个数学老师，给定一道数学问题，你需要判断学生答案和标准答案是否一致。如果学生的答案结果和标准答案结果一致，则得\end{CJK*}\\
\begin{CJK*}{UTF8}{gkai} 1分，如果不一致，则直接得0分。请按照"得分:"这样的形式输出学生分数。\end{CJK*}\\ 
\begin{CJK*}{UTF8}{gkai}You are a math teacher and you need to check if a student's answer to a math problem matches the standard answer. If the student's answer matches\end{CJK*}\\
\begin{CJK*}{UTF8}{gkai}  the standard answer, they receive 1 point. If not, they receive 0 points. Please output the student's score in the format of "Score:".\end{CJK*}} \\
\hline

Code  & \makecell[l]{ \begin{CJK*}{UTF8}{gkai} 你是一个计算机科学老师，给定一道编程问题，你需要判断学生答案是否能够顺利执行并取得满足题目要求的结果。如果可以，则得\end{CJK*}\\
\begin{CJK*}{UTF8}{gkai}1分，不可以则得0分。你可以参考标准答案中的代码。请按照"得分:"这样的形式输出学生分数。\end{CJK*} \\
\begin{CJK*}{UTF8}{gkai}You are a computer science teacher who needs to evaluate whether a student's programming answer can successfully execute and achieve the\end{CJK*}\\
\begin{CJK*}{UTF8}{gkai} desired result for a given problem. If it can, the student gets 1 point, otherwise they get 0 points. You can refer to the code in the standard answer.\end{CJK*}\\
\begin{CJK*}{UTF8}{gkai} Please output the student's score in the format of "score:".\end{CJK*}} \\
\hline

COT  & \makecell[l]{ \begin{CJK*}{UTF8}{gkai} 你是一个逻辑学家，给定一个问题，你需要判断模型回答是否在符合常识、逻辑的前提下，很好的回答了这个问题。如果模型回答符\end{CJK*}\\
\begin{CJK*}{UTF8}{gkai}合逻辑，则模型回答得1分，如果模型回答不符合逻辑，则得0分。你可以参考标准回答中的内容。请按照"得分\end{CJK*}\\
\begin{CJK*}{UTF8}{gkai}:"这样的形式输出分数。\end{CJK*} \\
\begin{CJK*}{UTF8}{gkai}You are a logician, and given a question, you need to determine whether the model's answer is logical and in accordance with common sense. \end{CJK*}\\
\begin{CJK*}{UTF8}{gkai}If the model's answer is logical, it will receive a score of 1, and if it is not logical, it will receive a score of 0. You can refer to the content of the\end{CJK*}\\
\begin{CJK*}{UTF8}{gkai}  standard answer. Please output the score in the format of "Score:".\end{CJK*}} \\
\hline

Classification  & \makecell[l]{ \begin{CJK*}{UTF8}{gkai} 你需要通过参考标准答案，来对模型的答案给出分数，满分为1分，最低分为0分。请按照"得分:"这样的形式输出分数。评价标准要求\end{CJK*}\\
\begin{CJK*}{UTF8}{gkai}分类结果越准确，分数越高。\end{CJK*} \\
\begin{CJK*}{UTF8}{gkai}You need to give a score to the model's answer based on the reference standard answer, with a maximum score of 1 and a minimum score of 0.\end{CJK*}\\
\begin{CJK*}{UTF8}{gkai} Please output the score in the format of "Score:". The evaluation criteria require that the more accurate the classification result, the higher the\end{CJK*}\\
\begin{CJK*}{UTF8}{gkai} score.\end{CJK*}} \\
\hline

Extraction  & \makecell[l]{ \begin{CJK*}{UTF8}{gkai} 你需要通过参考标准答案，来对模型的答案给出分数，满分为1分，最低分为0分。请按照"得分:"这样的形式输出分数。评价标准要求\end{CJK*}\\
\begin{CJK*}{UTF8}{gkai}需要保证抽取出来的结果来自文本，并且符合问题的要求。\end{CJK*} \\
\begin{CJK*}{UTF8}{gkai}You need to score the model's answer based on the reference standard answer, with a full score of 1 point and a minimum score of 0 point. Please\end{CJK*}\\
\begin{CJK*}{UTF8}{gkai} output the score in the format of "Score:". The evaluation criteria require that the extracted results come from the text and meet the requirements\end{CJK*}\\
\begin{CJK*}{UTF8}{gkai} of the question.\end{CJK*}} \\
\hline

Open QA  & \makecell[l]{ \begin{CJK*}{UTF8}{gkai} 你需要通过参考标准答案，来对模型的答案给出分数，满分为1分，最低分为0分。请按照"得分:"这样的形式输出分数。评价标准要求\end{CJK*}\\
\begin{CJK*}{UTF8}{gkai} 回答的结果越接近正确答案分数越高。\end{CJK*} \\
\begin{CJK*}{UTF8}{gkai}You need to score the model's answer by referring to the standard answer, with a maximum score of 1 and a minimum score of 0. Please output\end{CJK*}\\
\begin{CJK*}{UTF8}{gkai} the score in the format of "Score: ". The evaluation standard requires that the closer  the answer given is to the standard answer, the higher\end{CJK*}\\
\begin{CJK*}{UTF8}{gkai} the score.\end{CJK*}} \\
\hline

Closed QA  & \makecell[l]{ \begin{CJK*}{UTF8}{gkai} 你需要通过参考标准答案，来对模型的答案给出分数，满分为1分，最低分为0分。请按照"得分:"这样的形式输出分数。评价标准要\end{CJK*}\\
\begin{CJK*}{UTF8}{gkai} 求回答的结果准确，且回答结果来自问题里面提供的信息。\end{CJK*} \\
\begin{CJK*}{UTF8}{gkai}You need to score the model's answer by referencing the standard answer. The full score is 1 point, and the lowest score is 0 point. Please output\end{CJK*}\\
\begin{CJK*}{UTF8}{gkai} the score in the format of "Score:". The evaluation criteria require that the answer is accurate and comes from the information provided in the \end{CJK*}\\
\begin{CJK*}{UTF8}{gkai} question.\end{CJK*}} \\
\hline

Generation  & \makecell[l]{ \begin{CJK*}{UTF8}{gkai}假设你是一个作家,你需要研究评价标准来对模型的答案给出分数，满分为1分，最低分为0分。请按照"得分:"这样的形式输出分数。\end{CJK*}\\
\begin{CJK*}{UTF8}{gkai}评价标准要求生成的结果语句通顺，内容主题符合要求。\end{CJK*} \\
\begin{CJK*}{UTF8}{gkai}Assuming you are a writer, you need to research evaluation criteria to give a score to the model's answer, with a maximum score of 1 point and\end{CJK*}\\
\begin{CJK*}{UTF8}{gkai} a minimum score of 0 points. Please output the score in the format of "Score:". The evaluation criteria require the generated sentence to be\end{CJK*}\\
\begin{CJK*}{UTF8}{gkai} smooth and the content to be relevant to the topic.\end{CJK*}} \\
\hline

Brainstorming  & \makecell[l]{\begin{CJK*}{UTF8}{gkai}你需要研究评价标准来对模型的答案给出分数，满分为1分，最低分为0分。请按照"得分:"这样的形式输出分数。评价标准要求要求\end{CJK*}\\
\begin{CJK*}{UTF8}{gkai}回答的内容对于问题有帮助，并且是真实没有恶意的。\end{CJK*} \\
\begin{CJK*}{UTF8}{gkai}You need to study the evaluation criteria to give a score to the model's answer, with a maximum score of 1 point and a minimum score of 0\end{CJK*}\\
\begin{CJK*}{UTF8}{gkai} points. Please output the score in the format of "Score:". The evaluation criteria require that the answer is helpful to the question and is\end{CJK*}\\
\begin{CJK*}{UTF8}{gkai} truthful and non-malicious.\end{CJK*}} \\
\hline

Rewrite  & \makecell[l]{ \begin{CJK*}{UTF8}{gkai}假设你是一个作家,你需要研究评价标准来对模型的答案给出分数，满分为1分，最低分为0分。请按照"得分:"这样的形式输出分数 。\end{CJK*}\\
\begin{CJK*}{UTF8}{gkai}评价标准要求重写过后的句子保持原有的意思，并且重写过后的句子越通顺分数越高。\end{CJK*} \\
\begin{CJK*}{UTF8}{gkai}Assuming that you are a writer, you need to research the evaluation criteria to give a score for the model's answer, with a maximum score of 1\end{CJK*}\\
\begin{CJK*}{UTF8}{gkai} point and a minimum score of 0 points. Please output the score in the format of "Score:". The evaluation criteria require that the rewritten\end{CJK*}\\
\begin{CJK*}{UTF8}{gkai} sentence retains the original meaning, and the more fluent the rewritten sentence, the higher the score.\end{CJK*}} \\
\hline

Translation  & \makecell[l]{ \begin{CJK*}{UTF8}{gkai} 假设你是一个语言学家，你需要通过参考标准答案，来对模型的答案给出分数，满分为1分，最低分为0分。请按照"得分:"这样的形\end{CJK*}\\
\begin{CJK*}{UTF8}{gkai}式输出分数 。评价标准要求翻译过后的句子保持原有的意思，并且翻译过后的句子越通顺分数越高。\end{CJK*} \\
\begin{CJK*}{UTF8}{gkai}Assuming you are a linguist, you need to score the model's answer based on the reference answer, with a full score of 1 point and a minimum\end{CJK*}\\
\begin{CJK*}{UTF8}{gkai} score of 0 point. Please output the score in the form of "Score:". The evaluation criteria require that the translated sentence retains the original\end{CJK*}\\
\begin{CJK*}{UTF8}{gkai} meaning and the more fluent the translation, the higher the score. \end{CJK*}} \\
\hline

Summarization  & \makecell[l]{ \begin{CJK*}{UTF8}{gkai} 假设你是一个作家,你需要通过参考标准答案，来对模型的答案给出分数，满分为1分，最低分为0分。请按照"得分:"这样的形式输出\end{CJK*}\\
\begin{CJK*}{UTF8}{gkai}分数。评价标准要求生成的摘要内容能包含输入文本信息的重点.\end{CJK*} \\
\begin{CJK*}{UTF8}{gkai}Assuming you are a writer, you need to score the model's answer by referring to the standard answer, with a full score of 1 point and a minimum\end{CJK*}\\
\begin{CJK*}{UTF8}{gkai} score of 0 points. Please output the score in the form of "Score:" The evaluation criteria require that the generated summary content can contain\end{CJK*}\\
\begin{CJK*}{UTF8}{gkai} the key points of the input text. \end{CJK*}} \\
\hline
\end{tabular}}
\end{center}
\label{eval_prompts}
\small
\end{table*}

\end{document}